\documentclass[10pt,twocolumn,letterpaper]{article}

\usepackage{iccv}              

\definecolor{iccvblue}{rgb}{0.21,0.49,0.74}
\usepackage[pagebackref,breaklinks,colorlinks,allcolors=iccvblue]{hyperref}
\usepackage{tabularx}
\usepackage{graphicx} 
\usepackage{float}
\usepackage{placeins}
\def\paperID{11891} 
\def\confName{ICCV}
\def\confYear{2025}

\title{MirrorMe: Towards Realtime and High Fidelity \\ Audio-Driven Halfbody Animation}

\author{
    Dechao Meng\footnotemark[1] \quad Steven Xiao\footnotemark[1] \quad Xindi Zhang\footnotemark[1]  \quad Guangyuan Wang\footnotemark[2] \\ Peng Zhang \quad Qi Wang \quad Bang Zhang \quad Liefeng Bo \\
    Tongyi Lab, Alibaba Group  \\
    \tt\url{https://youtu.be/5RxGawDro3s}
}

\begin{document}
\maketitle


\renewcommand{\thefootnote}{\fnsymbol{footnote}}
\footnotetext[1]{Equal contribution.}
\footnotetext[2]{Project leader.}

\begin{abstract}
Audio-driven portrait animation, which synthesizes realistic videos from reference images using audio signals, faces significant challenges in real-time generation of high-fidelity, temporally coherent animations. While recent diffusion-based methods improve generation quality by integrating audio into denoising processes, their reliance on frame-by-frame UNet architectures introduces prohibitive latency and struggles with temporal consistency. This paper introduces MirrorMe, a real-time, controllable framework built on the LTX video model—a diffusion transformer that compresses video spatially and temporally for efficient latent space denoising. To address LTX’s trade-offs between compression and semantic fidelity, we propose three innovations: (1) A reference identity injection mechanism via VAE-encoded image concatenation and self-attention, ensuring identity consistency; (2) A causal audio encoder and adapter tailored to LTX’s temporal structure, enabling precise audio-expression synchronization; and (3) A progressive training strategy combining close-up facial training, half-body synthesis with facial masking, and hand pose integration for enhanced gesture control. Extensive experiments on the EMTD Benchmark demonstrate MirrorMe’s state-of-the-art performance in fidelity, lip-sync accuracy, and temporal stability.
\end{abstract}    
\section{Introduction}
\label{sec:intro}

Audio Driven portrait animation typically refers to the process of generating high-quality animated videos from a reference portrait by utilizing audio as the driving signal. Recent years, audio driven portrait animation has gain broad applications in various fields, including film and animation production, game development, and education. However, as the volume of the internet continues to expand, the demand for AI created portrait animation greatly grows, making the real-time generation of high-quality, highly accurate audio-driven portrait animation a significant challenge.

Previous works primarily utilized GANs~\cite{goodfellow2020generative} as the generative models to animate portrait images, and achieve efficient inference processes~\cite{guo2024liveportrait,liu2025contextual,xu2025vasa,zhang2023metaportrait,sun2023vividtalk,ren2021pirenderer,gao2023high}. However, GAN-base models exhibit fundamental limitations in synthesizing intricate foreground-background compositions, the generated videos often lack expressiveness, naturalness, and realism, particularly for audio-driven animations involving large angles, extensive movements, and foreground occlusions. 


\begin{figure}[t]
\begin{center}
\includegraphics[width=1\linewidth]{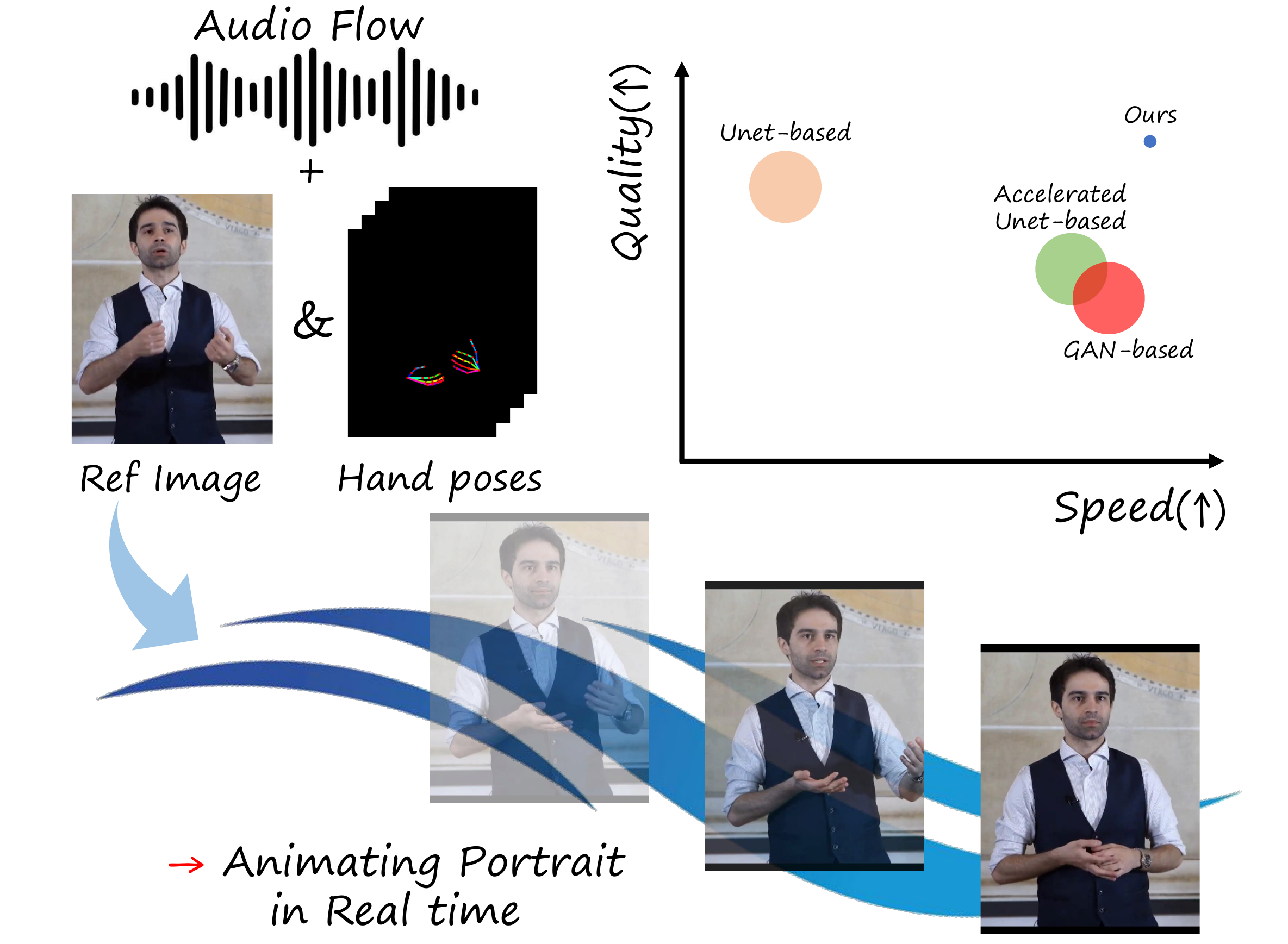}
\end{center}
\vspace{-1.5em}
   \caption{ MirrorMe is a DiT-base audio-driven portrait animation framework. Given a halfbody reference image and speech audio, it generates high-quality lip-synced videos in realtime.   
}
\vspace{-1em}
\label{fig:intro}
\end{figure}

Recent years, with the rapid advancement of diffusion UNets~\cite{blattmann2023stablevideodiffusionscaling,singer2022make,zhou2022magicvideo,guo2024animatediffanimatepersonalizedtexttoimage}, many works have achieved high-quality, realistic, and end-to-end audio driven portrait animation. They integrate audio signals into the denoising process, so as to generate videos that are highly consistent with audio sequences. These methods~\cite{shen2023difftalk,sun2023vividtalk,ma2023dreamtalk,tian2024emo,xu2024hallo} not only capture subtle facial expression changes, but also automatically generates lip movements and head poses that match the input audio signals. Despite the impressive generation performance of diffusion UNets, there still remain several challenges. First, the discrete sampling process introduces significant latency issues, where generating a few seconds of video may take over ten minutes. Second, these methods operate latent denoising at image level,
which result in poor temporal consistency among the generated frames. 
This inconsistency leads to flickering visual effects in the video, and greatly harms the generative fidelity of portrait animation. 



Building upon the aforementioned observations, this paper proposes a real-time, high-fidelity, and controllable portrait animation framework, namely \textbf{MirrorMe}. We adopt the LTX~\cite{hacohen2024ltxvideorealtimevideolatent} video model as our core architecture. Unlike UNet-base models that encode and denoise video frame by frame, the LTX model employs a diffusion transformer paradigm~\cite{cogvideox,polyak2024moviegen, kong2024hunyuanvideo,openai2024sora}, which compresses video from both spatial and temporal perspectives. It performs denoising steps in video latent space, which guarantees the temporal coherency of generative outcomes. While achieving real-time performance through exceptionally high spatio-temporal compression ratios, LTX's architecture inherently induces semantic degradation in both visual appearance and motion patterns. Such trade-off gives rise to significant challenges for audio-driven portrait animation, particularly regarding (a) identity consistency and (b) precise control signal adherence during denoising process.

To cope with the above challenges, this work develops a framework comprising three key innovations:
(1) As for identity preservation, we reuse the VAE from LTX to encode the reference image and concatenate it with noisy latents along the temporal dimension. The combined representation is then processed through self-attention to inject identity information. (2) To accommodate the highly compressed temporal encoding structure of LTX, we design a corresponding causal audio encoder and an audio adapter module. These components ensure that the expression changes driven by audio are naturally and accurately reflected in the generated video. (3) The high spatial-temporal compression of LTX model leads to considerable reduction of facial semantics, and this increases the difficulty of audio-driven facial expression control. To this end, we design a progressive training strategy: Initially training on close-up facial portraits, then extending to half-body synthesis while preserving audio-responsive facial dynamics through face masks. Additionally, hand pose signals are introduced to achieve precise gesture control.


We conducted comprehensive experimental evaluations to validate the effectiveness and real-time performance of MirrorMe. Our approach achieves state-of-the-art (SOTA) results across multiple critical metrics on the EMTD Benchmark~\cite{EchomimicV2}, significantly outperforming existing approaches and demonstrating its technical superiority. Furthermore, thorough comparative analyses were performed on temporal stability, identity consistency, and inference efficiency against competing methods. The experimental results substantiate that our framework can generate high-fidelity, controllable half-body animation under real-time constraints, while maintaining robust temporal coherence and identity preservation.

\begin{figure*}[t]
\begin{center}
\includegraphics[width=1\linewidth]{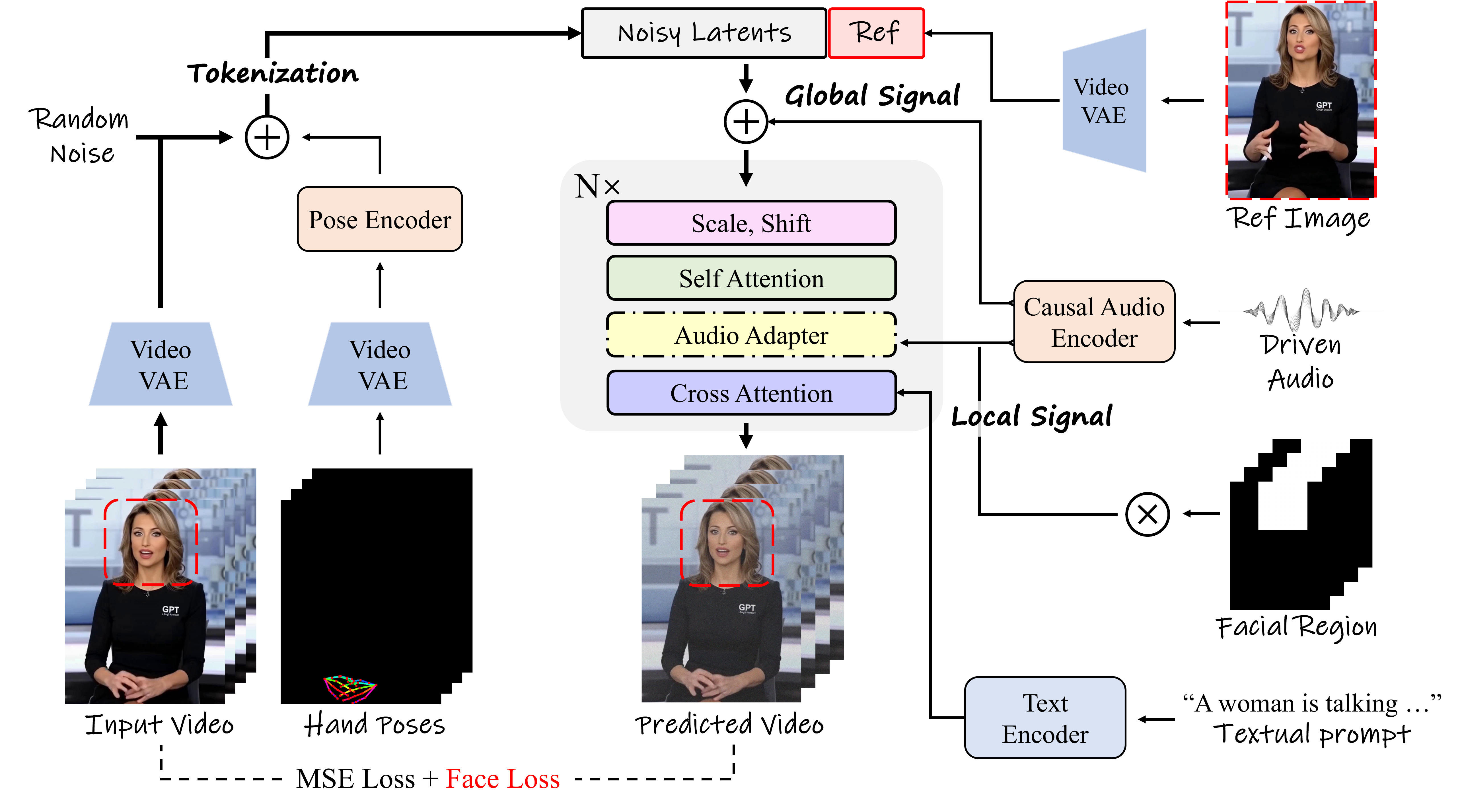}
\end{center}
\vspace{-1.5em}
   \caption{The overall framework of MirrorMe. 
   During training, the reference image is encoded through the VAE, and the encoded features are concatenated with the noisy latents. The audio input is encoded via the causal audio encoder and subsequently injected into the generative sequence through the audio adapter module. Facial regions are employed to spatially constrain the influence scope of audio signals. Concurrently, hand keypoints are integrated into the latents via the Pose Encoder to achieve precise hand pose control.
}
\vspace{-1em}
\label{fig:method}
\end{figure*}

\section{Related Works}
\label{sec:related}

\subsection{Video Generation}
In recent years, the advent of latent diffusion models has led to significant advancements in video generation algorithms. Stable Diffusion\cite{rombach2022highresolutionimagesynthesislatent} represents a notable text-to-image model that generates images with high quality and fidelity. Building upon this foundation, methods such as AnimateDiff\cite{guo2024animatediffanimatepersonalizedtexttoimage} and SVD\cite{blattmann2023stablevideodiffusionscaling} have extended image diffusion models to video generation models by incorporating temporal modules. These approaches still adhere to a 2D+1D structure, where spatial and temporal dimensions are treated separately.

With the development of the DiT (Diffusion-based Transformer) model and the emergence of 3D Variational Autoencoders (VAEs)~\citep{kingma2013vae,zheng2024opensorademocratizingefficientvideo}, several works including Sora\cite{openai2024sora}, EasyAnimate~\cite{xu2024easyanimate}, CogVideoX\cite{yang2024cogvideox}, HunyuanVideo \cite{kong2024hunyuanvideo} and LTX \cite{hacohen2024ltxvideorealtimevideolatent} have further evolved these models by transitioning from a 2D+1D structure to a fully integrated 3D structure. The 3D VAEs offer higher compression efficiency over time compared to their image-based counterparts, providing advantages in temporal stability.







\subsection{Controllable Portrait Animation}



Given a reference image of a person and control signals (such as skeletal points, audio, or 3DMM), controllable portrait animation aims at generating a video of the specific person under control signals. These methods can be categorized into two main types based on the underlying networks they employ: \textbf{GAN-based methods} and \textbf{Diffusion-based methods}.

\textbf{GAN-based} methods have been widely used for generating motion videos from reference images and control signals. These methods typically leverage the power of Generative Adversarial Networks (GANs) to synthesize realistic motions. 
They employ multimodal control signals to predict an motion field, which is then used to warp the features of the reference image. The warped features are subsequently fed into a Generative Adversarial Network (GAN)~\cite{goodfellow2020generative} for image synthesis. Commonly used control signals include facial keypoints~\cite{siarohin2019first, MRAA, guo2024liveportrait}, parameters of a 3D Morphable Model~\cite{doukas2021headgan, sun2023vividtalk, zhang2023sadtalker}, and one-dimensional latent representations~\cite{drobyshev2024emoportraits, xu2025vasa}, among others.
However, due to the inherent limitations of GANs in generating large-scale motions, these methods often struggle with significant performance degradation when the motion amplitude is large.

\textbf{Diffusion-based methods} have recently gained popularity due to their ability to generate high-quality and temporally stable videos. These methods typically use diffusion models to iteratively refine the generated frames, resulting in more realistic and coherent motions. 
Pose-driven methods~\cite{güler2018densepose, xu2023magicanimate, hu2023animate, ma2024follow} use explicit control signals like skeletal points or dense pose information to guide the motion generation process. These approaches are particularly effective for generating complex body movements.
Some methods~\cite{emo, jiang2024loopy, lin2024cyberhost, Vlogger, EchomimicV2} use audio signals to generate synchronized facial and body motions. 
These methods are capable of producing highly realistic talking head videos with accurate lip-syncing.
However, these methods often suffer from high computational costs and slow inference speeds. 
Furthermore, most of these approaches, which are fundamentally grounded in image-based model, frequently demonstrate compromised temporal coherence.

Our method is based on a lightweight video-based DiT (Diffusion Transformer) that significantly reduces computational complexity by compressing multiple frames temporally. This approach not only improves inference speed but also enhances temporal consistency. 

\section{Method}
\label{sec: Method}


The overall architecture of our model is illustrated in Figure~\ref{fig:method}. This section elaborates on the technical details of MirrorMe. Specifically, Section~\ref{subsec: preliminaries} presents the foundational preliminaries underlying our methodology, Section~\ref{subsec: appr_guid} details the reference-guided appearance integration mechanism, Section~\ref{subsec: audio_fuse} describes the audio-visual fusion module with temporal-spatial alignment, and Section~\ref{subsec: strategy} introduces the progressive training strategy for enhanced model convergence.

\subsection{Preliminaries}
\label{subsec: preliminaries}







Our method builds upon transformer-based latent diffusion models. Below, we outline the key components that underpin our approach.

\noindent\textbf{Transformer-Based Latent Diffusion Model}.
We adopt diffusion transformer ~\cite{peebles2023scalablediffusionmodelstransformers} as our basic model architecture. Given input video frames $\mathbf{F}\in\mathbb{R}^{T\times H\times W}$, we first encode $\mathbf{F}$ to the video latent $\mathbf{x_0}$ with a causal 3D-VAE $\mathcal{E}$:

\begin{equation}
\mathbf{x_0} = \mathcal{E}(\mathbf{F}),
\end{equation}

\noindent inspired by Rectified Flow \cite{lipman2023flowmatchinggenerativemodeling} and SD3 \cite{esser2024scalingrectifiedflowtransformers}, the clean input latent $ x_0 $ is linearly noised during the forward process according to:
\begin{equation}
\mathbf{x_t} = (1 - t)\mathbf{x_0} + t\mathbf{e},
\end{equation}
where $ t \in [0, 1] $ represents the timestep, and $ e $ is sampled from a standard-normal distribution $ \mathcal{N}(0, I) $. This process facilitates controlled noise introduction, enhancing the robustness of the generative pipeline.

Given the noisy latent $\mathbf{x}_t$ and timestep $t$, the model is trained to predict the velocity ${\rm \mathbf{u}}_t=d {\rm \bf{x}}_t/dt$, which guides the sample ${\rm \mathbf{x}}_t$ towards the sample ${\rm \mathbf{x}}_1$. 
The model parameters are optimized by minimizing the mean squared error between the predicted velocity ${\rm \mathbf{v}}_t$ and the ground truth velocity ${\rm \mathbf{u}}_t$, expressed as the loss function:
\begin{equation}
  \label{eq:fm}
  \mathcal{L}_{{\rm MSE}}=\mathbb{E}_{t,{\rm \bf{x}}_0,{\rm \bf{x}}_1}\|{\rm \bf{v}}_t - {\rm \bf{u}}_t  \|^2.
\end{equation}


\noindent\textbf{Lightweight Tokenization with LTX}.
Our framework utilizes the LTX architecture \cite{hacohen2024ltxvideorealtimevideolatent} that establishes a remarkable 1:8192 compression ratio via spatiotemporal tokenization ($ 32 \times 32 \times 8 $ pixels per token). 
This design enables real-time video synthesis, generating 5-second sequences at 24 fps ($ 768 \times 512 $  resolution) within 2 seconds on Nvidia H100 GPUs. 
Then it implements Rotary Position Embedding (RoPE) \cite{su2023roformerenhancedtransformerrotary} across height, width, and temporal (HWT) dimensions for each latent token. This multi-axis positional encoding mechanism enhances spatiotemporal coherence in video synthesis by establishing explicit geometric relationships between tokens across frames. 
However, such extreme compression introduces fundamental trade-offs: Although LTX demonstrates strong performance for static scenes with limited human presence and low camera dynamics, it exhibits limitations in capturing intricate human kinematics. The system shows degraded temporal coherence when processing high-amplitude motions, frequently manifesting as structural distortions and temporal discontinuities in articulated body movements.

\begin{figure}[t]
\begin{center}
\includegraphics[width=1\linewidth]{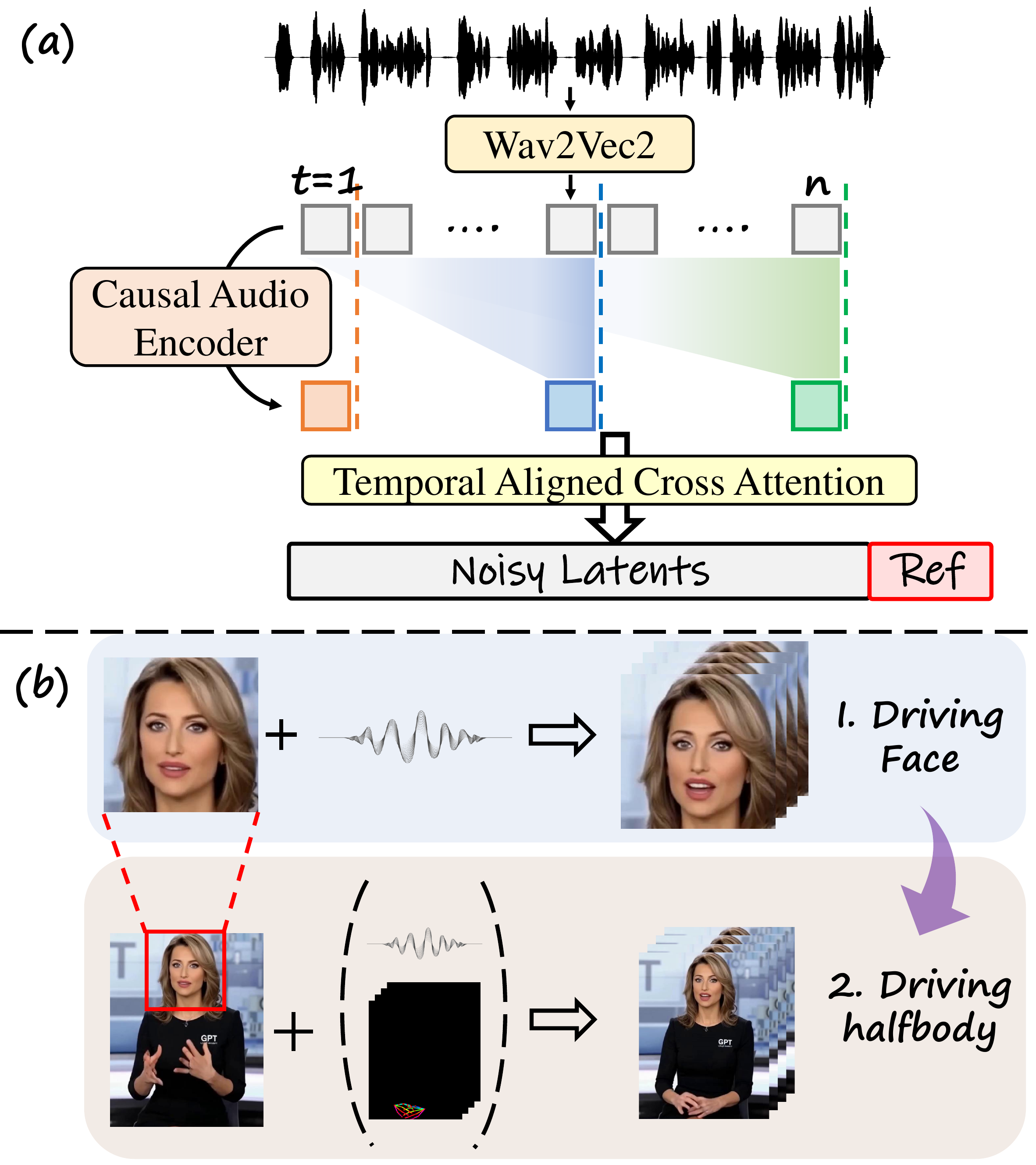}
\end{center}
\vspace{-1.5em}
   \caption{Illustration of (a) causal audio encoder and (b) progressive training pipeline.}
\vspace{-1em}
\label{fig:audio_encoder}
\end{figure}

To this end, we carefully designed our framework in a targeted manner, achieving an optimal balance among generation efficiency, output quality, and controllability.

\subsection{Appearance Integration}
\label{subsec: appr_guid}



Given a reference portrait image, we aim at integrating its appearance information into noisy latent sequences, so as to generate video sequences consistent with the reference identity (ID). To achieve this, we leverage the 3D-VAE from the LTX model to encode the reference portrait. The encoded reference latents are concatenated with the original noisy latents and fed into transformer blocks, enabling ID-aware information injection through self-attention mechanisms that mediate interactions between the generated video and the reference. 

Given the absence of inherent temporal relationships between reference images and video sequences, we modify the RoPE coordinates of reference latents during unified self-attention computation to maintain the LTX model's inherent temporal consistency. Specifically, we spatially align the reference latents with noisy latents while assigning a larger temporal offset to the reference latents. This design ensures effective injection of reference appearance information without disrupting the intrinsic temporal coherence across generated video frames.

To enable long-term video generation, we introduce motion frames during the synthesis process. During training, we randomly replace the initial frames of noisy latent sequences with ground truth video frames. At inference time, the final frames from preceding video clips are utilized as motion priors for subsequent clip generation, establishing persistent temporal anchors that facilitate coherent long-term video synthesis while maintaining motion consistency across segments.

\subsection{Audio Fusion Module}
\label{subsec: audio_fuse}

The LTX-based audio-driven video generation faces critical spatiotemporal misalignment between driving audio sequence and noisy video latents: (1) From the temporal perspective, the cross-modal misalignment arises from the non-uniform temporal compression in Causal 3D VAE (as illustrated in Figure~\ref{fig_03}(a)), which disrupts frame-level correspondence between visual and acoustic features. (2) From the spatial perspective, as one-dimensional signals, speech features primarily affect localized facial regions, inherently lack structural awareness. To address these challenges, we propose a causal audio encoder and an attention-based audio adapter, establishing precise cross-modal alignment for natural lip synchronization.




\noindent\textbf{Causal Audio Encoder}. To resolve the temporal incompatibility between the compressed visual representation and input audio features, we propose a causal audio encoder that achieves consistent temporal compression while strictly preserving temporal causality. 

The proposed audio processing pipeline is illustrated in Figure~\ref{fig:audio_encoder}(a). It is initiates with a pre-trained wav2vec2~\cite{baevski2020wav2vec20frameworkselfsupervised,schneider2019wav2vecunsupervisedpretrainingspeech}  model that extracts frame-level acoustic embeddings from raw audio signals. To establish temporal correspondence with visual features, we architect a causal audio encoder according to the hierarchical structure of video VAEs. Specifically, the module incorporates three stacked 1D causal convolutional downsampling layers, progressively compressing the audio sequence to match the temporal resolution of video latent representations. 


Given a noisy latent $\mathbf{x_t}\in\mathbb{R}^{(\tilde{T}\times \tilde{H}\times \tilde{W})\times C}$, where $\tilde{T}$ represents the compressed temporal dimension, and $\tilde{H}, \tilde{W}$ represent the compressed spatial dimension, $C$ represents the channel of video latents. Our causal audio encoder processes raw audio waveforms of length $T$ into temporally aligned global acoustic feature $\mathbf{a}_{\text{glo}}\in \mathbb{R}^{\tilde{T}\times C_a}$ and local features $\mathbf{a}_{\text{loc}}\in \mathbb{R}^{\tilde{T} \times n \times C_a}$, where $n$ indexes distinct semantic constituents within the audio features. Inspired by the approach of timestep embedding, we add $\mathbf{a}_{\text{glo}}$ to noisy latents $\mathbf{x_t}$ in a temporally-aligned manner to achieve global-level motion modulation. Additionally, we employ the audio adapter to inject local audio features $\mathbf{a}_{\text{loc}}$ into video features at intervals of every $k$ transformer blocks.


\noindent\textbf{Audio Adapter}. The above local audio features $\mathbf{a}_{\text{local}}$ are temporal aligned with the noisy latents. However, distinct spatial regions within frame-specific latents response differently to auditory signals. For instance, while lip movements and facial expressions require audio-informed synthesis, background regions should remain acoustically invariant. To address this, we design an audio adapter that facilitates networks to adaptively learn audio-visual correspondence with cross-attention. We introduce learnable padding tokens $\mathbf{p} \in \mathbb{R}^{\tilde{T} \times 1 \times C_a}$ concatenated with $\mathbf{a}_{\text{loc}}$ and obtain the augmented audio features $\hat{\mathbf{a}}_{\text{loc}}\in \mathbb{R}^{\tilde{T} \times (n + 1) \times C_a}$:

\begin{equation}
  \label{eq:padding}
      \hat{\mathbf{a}}_{\text{loc}}=\mathbf{a}_{\text{loc}} \oplus  \mathbf{p}, 
\end{equation}

the augmented audio features then interact with visual latent codes $\mathbf{x}_t^i$ at layer $i$ through cross-attention, the temporal dimension of $x_t^i$ is reshaped to batch dimension for temporal-aligned feature interaction: 

\vspace{-2mm}
\begin{align}
  \mathrm{CrossAttn}(\textbf{Q}, \textbf{K}, \textbf{V}) &= \mathrm{Softmax}(\frac{\textbf{Q}\textbf{K}^\top}{\sqrt{d}})\textbf{V}, \label{eq:CA} \\
  \textbf{Q}=\mathbf{x}_t^i\textbf{W}_q,\ \textbf{K}=\hat{\mathbf{a}}&_{\text{loc}}\textbf{W}_k,\ \textbf{V}=\hat{\mathbf{a}}_{\text{loc}}\textbf{W}_v, \label{eq:CA_new}
\end{align}

\noindent the noisy latents is then updated is a residual way:

\begin{equation}
    \label{eq:update} 
    \mathbf{x}_t^i \leftarrow \mathbf{x}_t^i+\gamma\cdot\mathrm{LayerNorm}(\mathrm{CrossAttn}(\mathbf{x}_t^i, \hat{\mathbf{a}}_{\text{loc}})). 
\end{equation}

In this way, we modulate the responsiveness of distinct semantic regions to audio signals while establishing precise acoustic-driven control over facial expression.



\subsection{Progressive Training Strategy}
\label{subsec: strategy}


Due to the minimal proportion of facial regions within half-body portrait frames and the inherent spatiotemporal compression characteristics of LTX architecture, the associated facial semantics suffer from degradation, which consequently amplifies the control difficulty of driven signals.  Furthermore, unlike pose signals that provide explicit spatial clues of bodily movements, audio signals primarily exert implicit influence on facial and lip articulations. To address these challenges, as shown in Figure~\ref{fig_03}(b), we propose a audio-centric progressive training strategy to facilitate rapid establishment of cross-modal semantic correlations between facial regions and audio driving signals, thereby enhancing the audio-driven control capability over portrait animations. 

\noindent\textbf{The audio-centric progress training} mainly consists of two phases: (1) \textit{Audio-driven facial animation}, 
we isolate the head region through spatial cropping, and conduct localized audio-driven learning of facial features at an elevated resolution. This hierarchical approach accelerates the learning of audio-facial correlations while preventing interference from unrelated body dynamics. (2) \textit{Audio-driven halfbody animation}, 
building upon the above hierarchical framework, we incrementally expand the synthesis scope to encompass upper-body regions while maintaining seamless audio-driven generation continuity. To enable explicit hand motion control, we integrate hand keypoint conditioning through pose encoder module. 

\noindent\textbf{Facial Region Constraint}. During training of audio-driven halfbody animation, in order to effectively preserve the audio-facial semantic correlations established in the first-stage, and eliminate undesirable audio influences on non-facial regions, we implement a dynamic facial mask to modulate the computed audio attention maps. This spatial constraint ensures exclusive audio-driven semantic activation within facial areas.

During inference, we employ an enlarged static facial mask for cross-modal feature interactions, and this empirically achieves two critical improvements: (1) maintaining natural head motion dynamics while preventing excessive motion blur, and (2) implicitly compensating for the LTX architecture's inherent limitations in modeling high-frequency motion patterns through spatial regularization.

\noindent\textbf{Face Resampling Loss}. To further enhance audio-driven facial motion fidelity and improve identity (ID) consistency, we introduce face resampling loss to facilitate model learning textural fidelity and spatiotemporal motion semantics in facial dynamics. 
Specifically, we measure the discrepancy between the model prediction 
and ground truth video latents at the facial regions with L2 Loss: 

\begin{equation}
  \label{eq:face_loss}
  \mathcal{L}_{{\rm face}}=\mathbb{E}_{t,{\rm \bf{x}}_0,{\rm \bf{x}}_1}\|{\mathbf{M}_{\rm{face}}\odot(\rm \bf{v}}_t - {\rm \bf{u}}_t)  \|^2,
\end{equation}

where $\mathbf{M}_{\rm{face}}$ is the extracted face masks. Combined with the noise prediction loss in $\mathcal{L}_{\rm{MSE}}$ in Eq.~\ref{eq:fm}, our final learning objective can be written as below:

\begin{equation}
  \label{eq:loss_total}
  \mathcal{L}_{\rm{total}}=\mathcal{L}_{\rm{MSE}}+\lambda\mathcal{L}_{\rm{face}},
\end{equation}

\section{Experiments}
\subsection{Experimental Setup}
\label{subsec: exp_setup}

\begin{figure}[t]
\begin{center}
\includegraphics[width=1\linewidth]{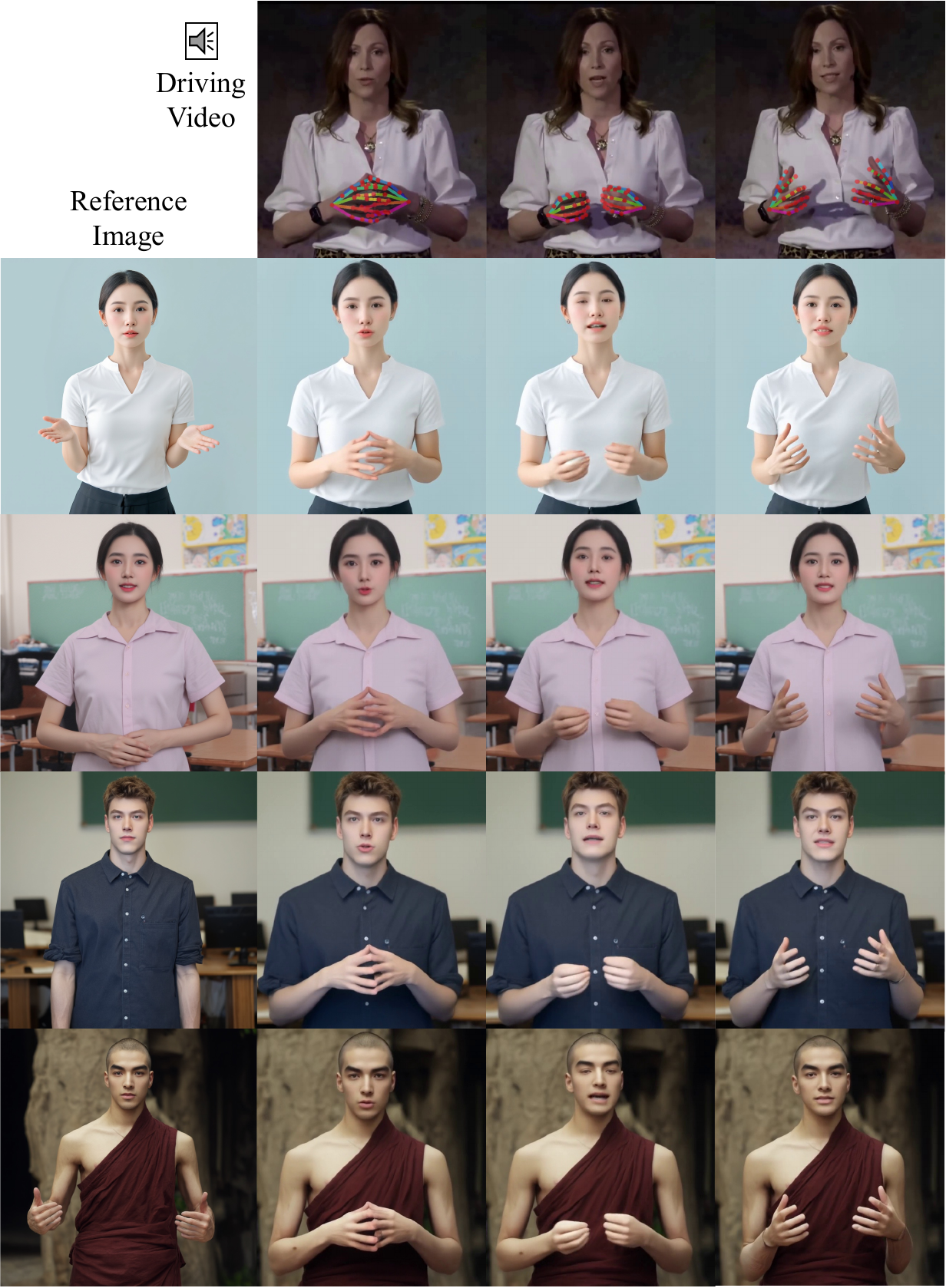}
\end{center}
\vspace{-1.5em}
   \caption{Cross-Identity Synthesis Results Driven by Multimodal Inputs (Audio/Hand Keypoints).
}

\vspace{-1em}
\label{fig:cross-id}
\end{figure}

\begin{figure*}[t!]
\begin{center}
\includegraphics[width=.95\linewidth]{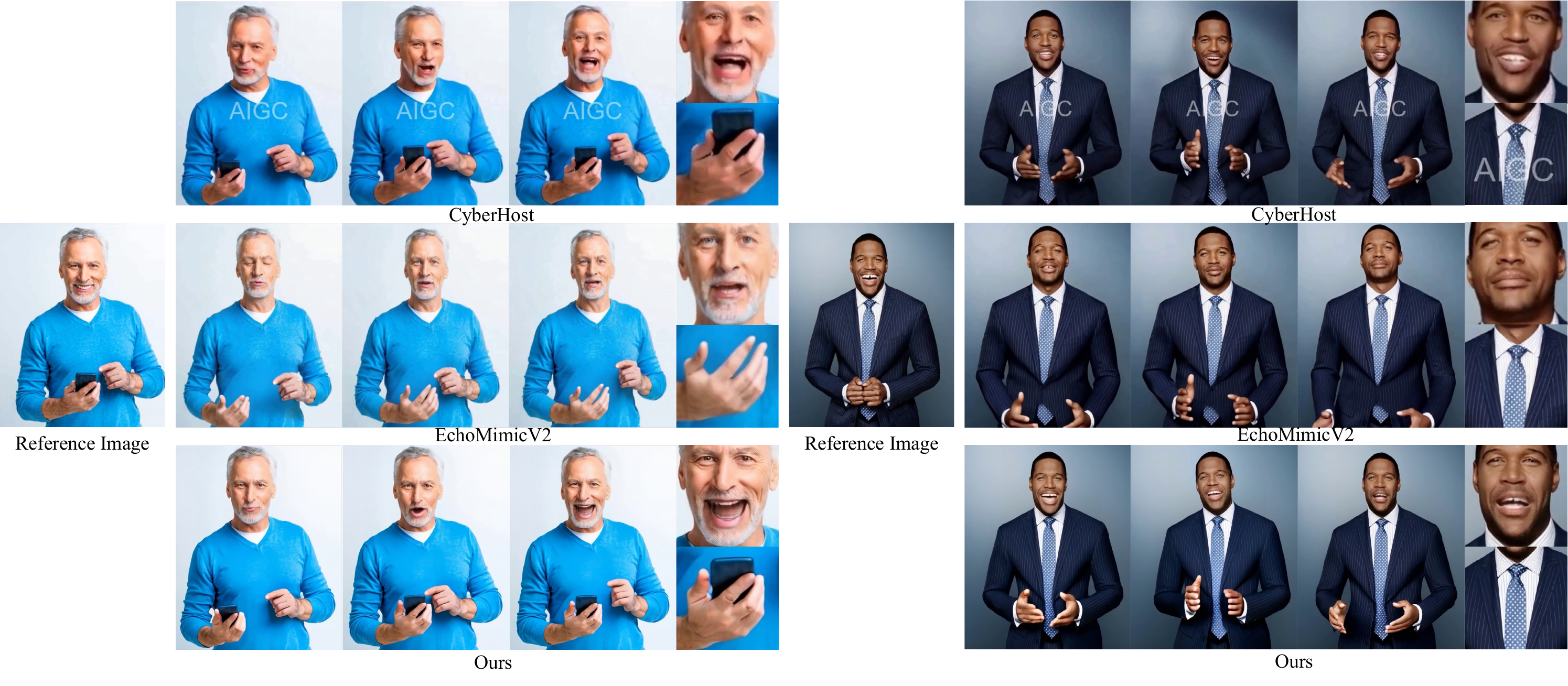}
\end{center}
\vspace{-1.5em}
   \caption{The qualitative results of MirrorMe compare to other audio-driven halfbody animation methods.
   Some local areas were enlarged for better comparison.
}
\vspace{-1em}
\label{fig_03}
\end{figure*}


\textbf{Training settings:} 
For optimization, we employ the Adam optimizer and conduct training across four NVIDIA A100 GPUs. The spatiotemporal resolution for each data is maintained at $121\times 512 \times 768$, with a batch size of 1 per GPU. The learning rate is consistently set at 4e-6 throughout the training process. To facilitate classifier guidance-free inference, we implement an independent dropout mechanism for different control signals with a probability of 10\%.
The first phase comprises 10,000 training steps, followed by an extended second phase of 30,000 steps, ensuring comprehensive model convergence and refinement.
To enhance the sampling probability of high-noised latents, we apply timestep shifting with a coefficient of 17 during training. 
This two-stage training strategy, combined with our carefully selected hyperparameters and augmentation techniques, enables the model to progressively learn complex audio-visual correlations while maintaining precise control over different body regions.


\textbf{Inference settings:} During the inference phase, we configure the classifier-free guidance scale to 2.5 and perform denoising over 15 steps. For unconditional inputs, we simultaneously zero out the reference image, audio signal, and hand pose data. 


\textbf{Dataset:} Due to the limited availability of open-source datasets specifically for upper-body digital human applications, we collected 4,053 distinct ID speech videos from YouTube. Following a meticulous scene segmentation process, we obtained a curated collection of 10,000 high-quality multilingual upper-body speech video clips, accumulating to a total duration exceeding 200 hours. This comprehensive dataset serves as our primary training resource, ensuring diverse representation of speech patterns and upper-body movements.

\subsection{Quantitative Comparisons}
\label{subsec: quan_comp}
For evaluation purposes, we use the EMTD\cite{EchomimicV2} dataset, which comprises 110 TED talk videos, as our benchmark testing dataset. This carefully selected dataset enables rigorous quantitative comparisons with existing methods, providing reliable metrics for assessing model performance across various dimensions of upper-body motion synthesis and speech synchronization.

\textbf{Evaluation Metrics.}
We conducted a systematic benchmarking analysis against state-of-the-art approaches using established quantitative metrics, including:
1) FID, FVD\cite{unterthiner2018towards}, SSIM\cite{wang2004image}, and PSNR\cite{hore2010image} are primarily employed to assess low-level visual quality. 
2) the cosine similarity (CSIM) between facial features of the reference image and the generated video frames is calculated to measure identity consistency. 
3) SyncNet\cite{prajwal2020lip} is utilized to compute Sync-C and Sync-D for validating the accuracy of audio-lip synchronization. 
4) For evaluating hand animation, the average Hand Keypoint Confidence (HKC) is adopted to assess hand quality in audio-driven scenarios, while Hand Keypoint Variance (HKV) is calculated to indicate the diversity of hand movements.

\textbf{Compared Methods.} 
We conducted comparative analysis with three diffusion based state-of-the-art upper-body digital human generation approaches
:
AnimateAnyone~\cite{hu2023animate} incorporates pose guidance through a lightweight PoseGuider adapter module in its diffusion architecture;
MimicMotion~\cite{mimicmotion2024} extends expression control by integrating facial landmarks;
and EchoMimicV2~\cite{EchomimicV2} introduces audio attention to control expression implicitly.




The comparison is shown in Table~\ref{tab: compare_sota}. 
Our method demonstrates substantial improvements across key video quality metrics, achieving state-of-the-art performance in SSIM, FID and FVD. These quantitative improvements confirm that our synthesized videos better preserve the original data distribution characteristics compared to existing approaches.
While conventional PSNR metrics – typically measuring pixel-level reconstruction fidelity – show a performance gap against pose-controlled methods.
This is because our method does not restrict pose and cannot achieve pixel-level alignment with the original data.
Beneficial from the progressive training strategy, our models perform better in lip synchronization(Sync-C, Sync-D) and identity preservation (CSIM) . 
Furthermore, the proposed hand guidance module achieves competitive results in hand kinematic metrics (HKC, HKV).

\subsection{Ablation Study}

\label{subsec: ablation}


We analyzed the contributions of various modules to the overall performance improvement. 
For the convenience of comparison, we have implemented a simple baseline model, where the audio features are injected into the LTX model through cross attention directly. On basis of this, we sequentially add casual audio fusion module, progressive training strategy and hand guider. 

As shown in Table~\ref{tab: compare_sota}, the baseline model fails to effectively capture audio information, resulting in poor SyncNet metrics.
As illustrated in the $5^{\rm{th}}$ row of Table~\ref{tab: compare_sota}, the incorporation of our designed audio module significantly enhances lip-sync accuracy. By integrating an advanced audio encoder and adapter, the model is better equipped to learn and utilize audio information, leading to notable improvements in lip synchronization.

Our progressive training strategy, which initially focuses on facial regions before extending to the upper body, further boosts SyncNet metrics. This approach allows the audio features to concentrate more effectively on the facial area, decoupling them from body posture. Consequently, this not only improves lip-sync accuracy but also enhances CSIM in the upper body scenarios by constraining the facial region.
The hand guider enables controlled generation of hand movements, leading to substantial improvements in HKV.

In summary, our comprehensive approach, which includes advanced audio encoding, progressive training strategies, and a dedicated hand guider module, results in significant enhancements across multiple metrics. These improvements are reflected in better lip-sync accuracy, enhanced identity consistency, and more realistic hand movements, thereby advancing the state-of-the-art in multimodal synthesis.

\begin{table*}[t]
    \begin{center}
    \resizebox{.97\textwidth}{!}{
    \begin{tabular}{l|cccc|cc|cc|c}
    \hline
    Methods & FID$\downarrow$ & FVD$\downarrow$ & SSIM$\uparrow$ & PSNR$\uparrow$ & Sync-D$\downarrow$ & Sync-C$\uparrow$ & HKC$\uparrow$ & HKV$\uparrow$ &CSIM$\uparrow$ \\
    \hline\hline
    AnimateAnyone\cite{hu2023animate}       &  58.98            & 1016.47          & 0.729          & 20.579          & 13.887           &   0.987           &   0.809           & 23.87           &  0.387 \\
    MimicMotion\cite{mimicmotion2024}  &  53.47            & 622.62           & 0.702          & 19.278          & 7.958            &   1.495           &   0.907           &  24.82          &  0.526 \\
    EchoMimicV2\cite{EchomimicV2}           &  49.33            & 598.45           & 0.738          & \textbf{21.986} & 7.021            &   7.219           &   \textbf{0.923}  & 25.28           &  0.558 \\
    \hline
    \hline
    Baseline                                &  62.38            & 973.87           & 0.689          & 16.645          & 13.285           &   1.207           &   0.697           & 8.89            & 0.610 \\
    +Audio Encoder \& Adapter               &  55.23            & 737.23           & 0.713          & 17.065          & 8.531            &   6.820           &   0.628           & 7.38            & 0.633 \\
    +Progressive Training                   &  50.88            & 654.72           & 0.738          & 18.328          & 6.972            &   7.014           &   0.619           & 8.33            & 0.717 \\
    +Hand Guider (MirrorMe)                            &  \textbf{43.53}   & \textbf{358.21}  & \textbf{0.776} & 19.506          & \textbf{6.986}   &   \textbf{7.232}  &   0.913           & \textbf{25.32}           & \textbf{0.719} \\
    \hline
    \end{tabular}
    }
    \end{center}
    \vspace{-1.5em}
    \caption{Quantitative comparison and ablation  study of our proposed MirrorMe and other SOTA methods.}
    \vspace{-1.1em}
    \label{tab: compare_sota}
\end{table*}    

\subsection{Qualitative Comparisons}
\label{subsec: vis_comp}

To obtain qualitative results, we utilized the demos available on CyberHost's~\cite{lin2024cyberhost} project page as test cases. We compared the performance of our method against both CyberHost and EchomimicV2. As illustrated in Figure~\ref{fig_03}, our approach demonstrates superior identity consistency, not only in facial and hand regions but also in preserving intricate details(\eg, cell phone and tie pattern).


Moreover, our method exhibits improved lip-sync accuracy. It performs particularly well in rendering teeth and oral structures, providing a more realistic and precise representation of mouth movements during speech. This enhanced precision contributes to a more natural and visually coherent output.

Figure~\ref{fig:cross-id} demonstrates cross-identity synthesis capabilities, where reference images with diverse poses and anthropometric variations are driven by same audio signals and hand motion keypoints extracted from the source video. Our method achieves simultaneous optimization of lip-sync accuracy and precise hand gesture, while maintaining identity preservation alignment across all test cases.
\begin{figure}[b]
\begin{center}
\includegraphics[width=1\linewidth]{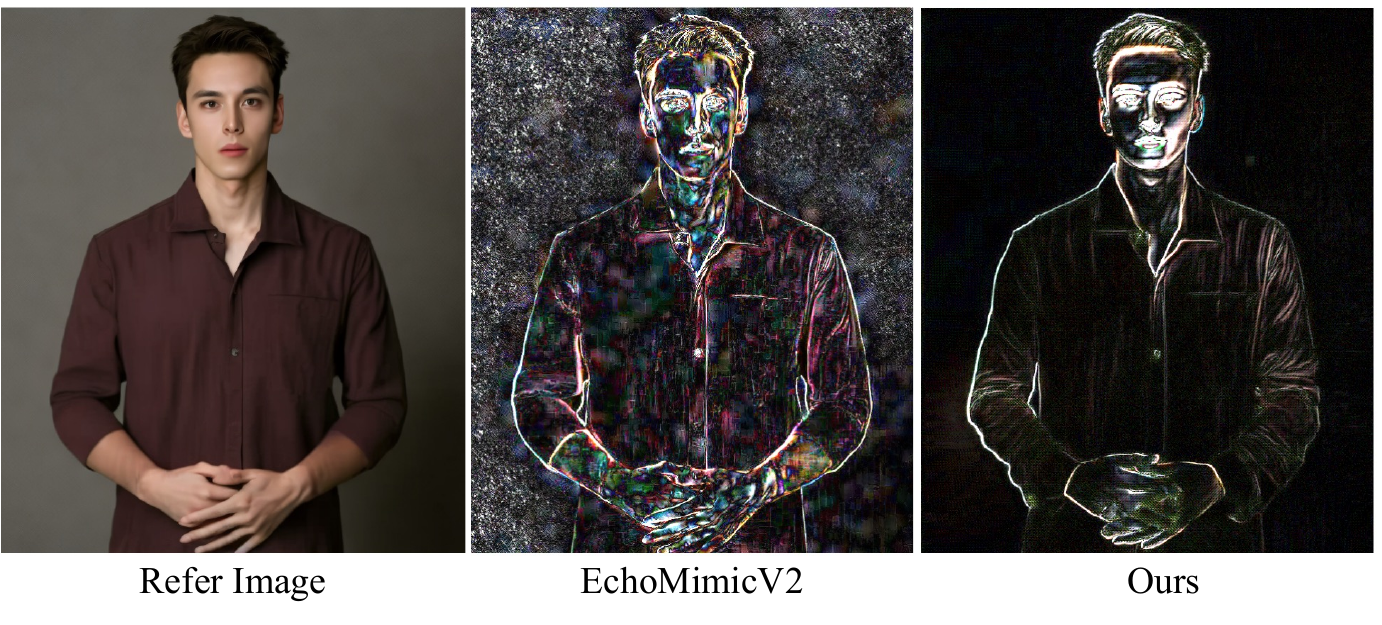}
\end{center}
\vspace{-1.5em}
   \caption{Comparison of the temporal consistency of our model and EchoMimicV2. For better visualization, the color scale have been magnified. 
}
\vspace{-1em}
\label{fig:time_consistency}
\end{figure}

\subsection{Temporal Consistency}
During practical testing, we observed that image-based methods often exhibit significant background flickering issues. This phenomenon arises because such methods primarily focus on optimizing single-frame generation quality, while insufficiently modeling temporal continuity in videos. Leveraging the strengths of the DiT architecture, our approach significantly enhances inter-frame stability. Figure~\ref{fig:time_consistency} illustrates a comparative analysis between our method and EchoMimicV2. We extracted two consecutive frames from the video and computed their difference. As shown in the figure, the difference map of EchoMimicV2 contains substantial noise, whereas our method maintains a nearly static background except for continuous variations around the human subject.

\subsection{Inference Efficiency}

\begin{table}[t]
\centering
\begin{tabular}{@{}llrrr@{}}
\toprule
Method         & Resolution       &  FPS  \\ \midrule
EchoMimicV2    & 768x768    & 0.32 \\
EchoMimicV2(Distilled)   & 768x768   & 3.20  \\
MirrorMe          & 768x512    & 24.33 \\ \bottomrule

\end{tabular}
\caption{Inference speed comparison with EchoMimic2. 
}
\vspace{-5mm}
\label{tab:speed}
\end{table}

Our method adopts the spatiotemporal compression strategy from LTX, which compresses the input video temporally to 1/8 of its original length and spatially to 1/32 of its original resolution in both height and width. This significantly reduces the number of latents that need to be processed by the attention mechanism. Given that the computational complexity of attention scales quadratically with the number of latents ($O(n^2)$), this configuration greatly enhances inference speed. 
Our framework demonstrates real-time generation capability at 512×768 resolution, achieving a rendering speed of 24 FPS on consumer-grade NVIDIA GPUs. This computational efficiency enables high-fidelity video output while maintaining temporal coherence, fulfilling the strict latency requirements for interactive digital human applications. Detailed comparison results of inference efficiency is shown in Table~\ref{tab:speed}.
Compared to UNet-based methods, 
our approach achieves nearly a 8x speedup at the same inference resolution and diffusion steps. 

\section{Conclusion}

In this paper, we address the critical challenges of real-time generation, identity preservation, and precise audio-driven control in high-fidelity portrait animation. Building upon the lightweight video diffusion architecture LTX, we carefully design an audio driven portrait animation framework, namely MirrorMe, and attain optimal balance among generation efficiency, output quality, and controllability.
Extensive experiments on the EMTD Benchmark demonstrate that MirrorMe achieves state-of-the-art performance in lip synchronization, expression fidelity, and temporal stability. 
This work bridges the gap between real-time inference and high-quality audio-driven animation, offering practical value for applications requiring interactive responsiveness. 


{
    \small
    \bibliographystyle{ieeenat_fullname}
    \bibliography{example_paper}
}

\end{document}